\documentclass[conference]{IEEEtran}
\usepackage{booktabs}
\usepackage{cite}
\usepackage{amsmath,amssymb,amsfonts}
\usepackage{algorithmic}
\usepackage{graphicx}
\usepackage{textcomp}
\usepackage{xcolor}
\usepackage{multirow}
\usepackage{makecell}
\usepackage{pdfpages}
\usepackage{svg}
\usepackage{url}
\usepackage{algorithm2e}
\usepackage[colorlinks,urlcolor=black, allcolors=blue]{hyperref}
\usepackage{fancyhdr}  % Added fancyhdr package

\def\BibTeX{{\rm B\kern-.05em{\sc i\kern-.025em b}\kern-.08em
    T\kern-.1667em\lower.7ex\hbox{E}\kern-.125emX}}

\def\bng{\bngx}

\font\bngx=bang10

\def\*#1*#2{o\null{#2}{#1}}

\def\sh#1{\setbox0=\hbox{#1}%
     \kern-.02em\copy0\kern-\wd0
     \kern.04em\copy0\kern-\wd0
     \kern-.02em\raise.0433em\box0 }
\begin{document}

\title{Detecting Racist Text in Bengali: An Ensemble Deep Learning Framework}

\fancypagestyle{firststyle}{
   \fancyhf{} % Clear header and footer
   \fancyhead[L]{\footnotesize Accepted and Presented in "2023 26th International Conference on Computer and Information Technology (ICCIT)"\\
   13-15 December, Cox’s Bazar, Bangladesh}

   \fancyfoot[L]{\fbox{\parbox{\textwidth}{\textcopyright 2023 IEEE. Personal use of this material is permitted.
  Permission from IEEE must be obtained for all other uses, in any current or future
  media, including reprinting/republishing this material for advertising or promotional
  purposes, creating new collective works, for resale or redistribution to servers or lists, or reuse of any copyrighted component of this work in other works.}}}

   \renewcommand{\footrulewidth}{0pt} 
   \renewcommand{\headrulewidth}{0pt} 
}

  % \fancyfoot[C]{\fbox{\parbox{\textwidth}{ \textcopyright 2023 IEEE. Personal use of this material is permitted.
  % Permission from IEEE must be obtained for all other uses, in any current or future
  % media, including reprinting/republishing this material for advertising or promotional
  % purposes, creating new collective works, for resale or redistribution to servers or lists, or reuse of any copyrighted component of this work in other works.}}}

\author{\IEEEauthorblockN{S. S. Saruar Jahan, Nusrat Jahan, Sadia Rahman Priota}
\IEEEauthorblockA{\textit{Dept of Computer Science \& Engineering} \\
\textit{Ahsanullah University of Science \& Technology}\\
Dhaka, Bangladesh \\
Email: sjalim71@gmail.com, itsnusratjahan19@gmail.com, sadiarahmanpriyota@gmail.com}
}

\maketitle
\thispagestyle{firststyle}

\begin{abstract}
Racism is an alarming phenomenon in our country as well as all over the world. Every day we have come across some racist comments in our daily life and virtual life. Though we can eradicate this racism from virtual life (such as Social Media). In this paper, we have tried to detect those racist comments with NLP and deep learning techniques. We have built a novel dataset in the Bengali Language. Further, we annotated the dataset and conducted data label validation. After extensive utilization of deep learning methodologies, we have successfully achieved text detection with an impressive accuracy rate of 87.94\% using the Ensemble approach. We have applied  RNN and LSTM models using BERT Embeddings. However, the MCNN-LSTM model performed highest among all those models. Lastly, the Ensemble approach has been followed to combine all the model results to increase overall performance.

\end{abstract}

\section{Introduction}
Racism, a pervasive social issue with roots dating back to the 19\textsuperscript{th} century, is fueled by human possessiveness concerning the concept of \emph{race}. It's important to note that \emph{race} is acknowledged as a social construct, representing the core values of a society that dictate human interactions. \emph{Racism} is defined as a perspective asserting the intrinsic superiority of a specific race, with race being considered a fundamental predictor of human characteristics and abilities. Various aspects contribute to \emph{racism}, including color, culture, economics, gender, ideology, and more.
The rise of social media has intensified the proliferation of this harmful mindset. The virtual realm allows for significant impact, with just a few words having the potential to shatter someone's life. \emph{Racism} is a specific form of hate speech, and sociologists globally classify it into various types. \emph{Racism} manifests based on factors such as race, culture, appearance, and social status. Individuals worldwide similarly encounter this phenomenon.
To address this concern, our research focuses on developing effective methods for detecting racist texts in Bengali social media comments. We have explored the application of deep neural network architectures for this task and demonstrated their efficacy in identifying instances of \emph{racism}. Several works \cite{SHARIF2022462,10101588} have been done in this domain earlier.

In this paper, our vision is to eradicate \emph{racism} from social media by detecting racist texts. Natural Language Processing (NLP) techniques have been used for text detection. RNN \& LSTM-based neural networks have been built to classify racism. We have tried to build a benchmark dataset from social media texts. We have pre-processed our text dataset by traditional NLP pre-processing techniques. \emph{BERT} pre-trained models have been for word embeddings. We have used \emph{sentence-transformer} for embeddings. Some studies \cite{10236810,10055276} can also be found that employed BERT models in Bengali classification tasks.

The study's results are encouraging; it was able to categorize a dataset in the Bengali language with more than 80\% accuracy into two binary classes: "Racism" and "Non-Racism." RNN and LSTM models were used, and the effect of pre-trained BERT models on results was seen. The study aims to make a substantial contribution to the fight against racism in the digital world, specifically in the Bengali language. The results are meant to provide guidance for initiatives and tactics aimed at reducing racism on social media. The study focuses on natural language processing (NLP) in text categorization. It builds a high-quality corpus from many social media platforms, systematically identifies racist text in Bengali, and compares standard and hybrid deep learning methods in-depth to evaluate their effectiveness.

\section{Related Works}
To find out more, we read a variety of pertinent literature on the same subjects. This made conducting this research easy for us. A quick review of the papers discussed below is provided.

%Paper1
The study\cite{reddy2023racism} employs the Stacked Ensemble GCR-NN Model for identifying racism in tweets. Sentiment analysis is conducted on 169,999 tweets, with the GRU at the top, CNN in the middle, and RNN at the bottom. The GCR-NN model demonstrates an impressive average accuracy of 98\%, surpassing the performance of traditional machine learning models. In contrast, SVM and LR can only detect 96\% and 95\% of racist tweets, respectively, while the GCR-NN model can identify 97\% of racist tweets with a 33\% misclassification rate.

%Paper2
The purpose of this study\cite{benitez2022detecting} is to use deep learning models on Twitter data to detect racism and xenophobia. The dataset was created using the Tweepy package. Five prediction models were developed, including two transfer learning models and three deep learning models. The best model achieved a f1 score of 85.76\% for non-racist, 84.52\% for racist, and 85.14\% for macro-averaged using the BETO model.
%Paper3
The research paper \cite{10101588} uses Deep Learning to classify Interpretable Multi Labeled Bengali Toxic Comments. The method uses LSTM networks, MConv-LSTM, and Bangla BERT embedding for binary classification and a combination of CNN and Bidirectional LSTM models for multi-label classification. The dataset includes 16,073 instances of toxic comments. The method captures complex structures and improves classification accuracy for Bengali toxicological observations, with LSTM networks and CNNBiLSTM models showing 89.42\% and 78.92\% accuracy respectively.
%Paper4
This paper\cite{karim2022multimodal} outlines a multimodal approach for Bengali hate speech detection using textual and visual information. It employs advanced neural architectures, including Bi-LSTM/Conv-LSTM, ConvNets, and multimodal fusion techniques. The largest dataset includes 4,500 classified memes. Conv-LSTM and XLM-RoBERTa models perform best for texts, while ResNet-152 and DenseNet-161 excel for memes. The study emphasizes feature selection and textual information's usefulness in hate speech identification, with memes contributing moderately.
%Paper5
This paper\cite{SHARIF2022462} presents a Bengali aggressive text dataset (BAD) using weighted ensemble techniques, including m-BERT, distil-BERT, Bangla-BERT, and XLM-R. The dataset, containing 14158 texts from social media sources, is categorized into religious, political, verbal, and gendered aggression classes. The dataset achieved the highest weighted f 1-score of 93.43\% in identification and 93.11\% in categorization tasks. Various techniques, including machine learning methods, deep learning techniques, and transformer-based models, were used to extract relevant features.\\
%Paper6-11
% Similarly, \cite{alotaibi2020racism, badjatiya2017deep, 10.14569/IJACSA.2020.0110861, 10.4102/td.v14i1.564,9666550, 10.3390/app10186527} various research uncovers indications of racism in a variety of languages. With the goal to locate racism in Arabic tweets, hate speech on Twitter, sentiment on Twitter feeds relating to the Afrikaner community of Orania in South Africa, and hate speech in Bangla videos, the study uses machine learning, deep learning, lexical analysis, and text mining techniques.
\section{Definition of the task}
We have thoroughly researched the racism definition regarding the Bengali language. Worldwide several types of racism exist. Some of those types are extinct. Considering the definition of several research works \cite{miles2004racism, braveman2022systemic, schmid1996definition, banton2015racism} and online platforms\footnote{\url{https://www.thoughtco.com/racism-definition-3026511}}.
We have come across three classes of Racism which are mostly found in social media text in the Bengali language which are \textbf{1)} Representational Racism, \textbf{2)} Ideological Racism, and \textbf{3)} Discursive Racism. All the types definitions are exhibited in Table \autoref{tab:typeDefinition}.
\begin{table*}[b]
    \centering
    \caption{Definitions of racism types in Bengali Language}
    \begin{tabular}{rp{14cm}}
    \hline
    \textbf{Type} & \textbf{Definition}\\
    \Xhline{3\arrayrulewidth}

 \rule{0pt}{10pt}Representational Racism & This type of racism, also known as symbolic or imagery racism, manifests itself through media, the arts, and other cultural manifestations. It entails continuing stereotypical images and biased representations of racial or ethnic groups, which can support prejudiced attitudes and views.\\
 \rule{0pt}{10pt}Ideological Racism & This refers to beliefs or ideologies that contend that some racial or ethnic groupings are inherently superior or inferior. It frequently has a strong hold on cultural standards, influencing attitudes and actions toward certain people.\\
 \rule{0pt}{10pt}Discursive Racism & Racist ideas and attitudes are sustained by discursive racism, which uses language, rhetoric, and discourse. Racial prejudices may be expressed directly or implicitly through public discourse, political debates, and everyday conversations.\\
        
    \Xhline{3\arrayrulewidth}
    \end{tabular}
    \label{tab:typeDefinition}
\end{table*}

\section{Data Acquisition}

\subsection{Data Collection}
Racism is a sub-domain of hate speech. We have collected 6k raw data from renowned social media platforms such as (FACEBOOK, YOUTUBE, and TWITTER). All the data were collected from public posts\footnote{\url{https://shorturl.ac/7azg3}}. There are quite a few corpora that are open-sourced. But those are mixed with different types of hate speech data. We intend to classify all the kinds of racism currently available in the Bengali language.\\
\subsection{Annotation Process}
We annotated the dataset for the multiclass target. But we have found mostly \textbf{three types} of racism altogether in the Bengali Language. Our dataset \& Code on GitHub\footnote{\url{https://github.com/sjalim/Detecting-Racist-Text-in-Bengali-An-\\Ensemble-Deep-Learning-Framework}} is open-sourced.
According to the definitions of \autoref{tab:typeDefinition}, we have annotated the data manually.
After that, we surveyed\footnote{\url{https://forms.gle/QNNE4TUqKkVX8ypS6}} that annotated data and corrected them accordingly.\\

\begin{figure} [h!]

% \includepdf[pages={{},-}, scale=0.4, fitpaper=true]{images/survey.pdf}
\includegraphics[scale=.4]{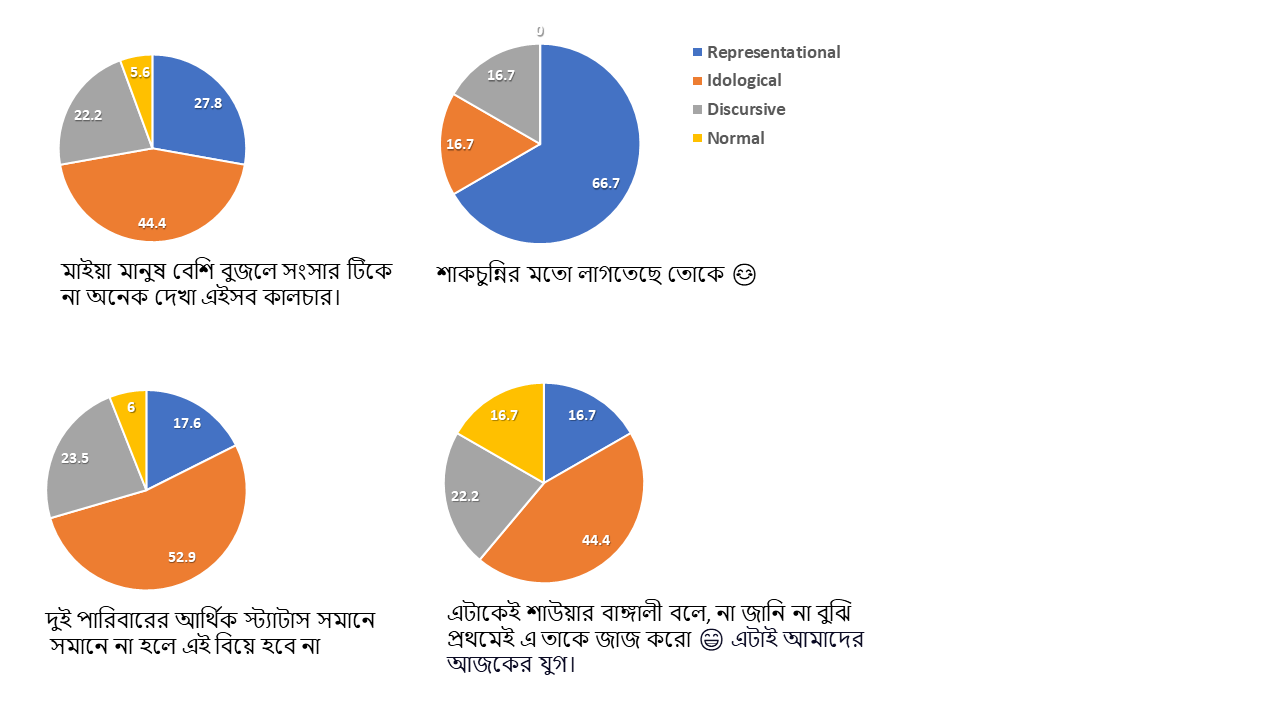}
\caption{Data Survey Report}
\label{fig:dataSurvey}
\end{figure}

\begin{table*}[]
\centering
\caption{Dataset Preview}
\begin{tabular}{p{4cm}p{4cm}p{7cm}}
\hline
\textbf{Text} & \textbf{Label} & \textbf{Remarks} \\
\hline
% {\bng আপনি একটা ঘোড়ার ডিম।}
{\bng Aapin EkTa egharhar iDm.}
& Representational Racism & Here, a person is judged for his appearance and nature with a horse egg which doesn't even exist. \\
\vspace{-16pt}
(You are a horse egg) && \\

\hline

% {\bng  মেয়েদের ইনকাম করা একটি বিশাল সামাজিক বিস্ফোরণ।} 
{\bng emJeedr {I}nkam kra EkiT ibshal samaijk ibes/pharN.}
& Ideological Racism & Some society think of an earning woman become a very troublesome. \\

% \vspace{-40pt}
(Earning girls is a huge social explosion) && \\
\hline

% {\bng  বস্তি বেডি একটা।} 
{\bng bis/t ebiD EkTa.}
& Discursive Racism & Here, the statement is denigrating women because she lives in a slum. \\
\vspace{-16pt}
(A slum woman) && \\

\hline
\end{tabular}
% \label{tab:typeDefinition}
\end{table*}

\subsection{Data Preprocessing}
\begin{itemize}

\item Number Remove :
A social media text contains a considerable number of values. Unfortunately, the context or information about the content provided by these numerical values is lacking. We have eliminated numerical values from our dataset for the sake of convenience.\\ Example: 
% {\bng ১২টার পরে সব রাস্তায় বস্তির পোলাপান থাকে।} 
\\ {\bng 12Tar per sb ras/tay bis/tr epalapan thaek.}
(Raw text) 
% {\bng টার পরে সব রাস্তায় বস্তির পোলাপান থাকে।}
\\ {\bng Tar per sb ras/tay bis/tr epalapan thaek.}
(Processed Text)
\item Punctuation Remove:
Since punctuation is not needed in sentences, removing it is a crucial NLP preprocessing step.\\ Example: 
% {\bng  বোকারাম!!!!! দূরে গিয়া মর.....।} 
\\{\bng ebakaram!!!!! duuer igya mr......|}
(Raw Text) 
% {\bng  বোকারাম দূরে গিয়া মর}
\\{\bng ebakaram duuer igya mr}
(Processed Text)
\item POS \footnote{\url{https://bnlp.readthedocs.io/en/latest/}} Remove :
Certain weakly dominant components of speech, including conjunctions, pronouns, interjections, and nouns, are present in our sample. We categorized the parts of speech for each word. \\Noun: 
% {\bng  রহিম, করিম}
{\bng riHm, kirm}
\\Pronoun: 
% {\bng আমি, তুমি, তোমার, তার }
{\bng Aaim, tuim, etamar, tar}
\\Conjunction: 
% \bn এবং, ও, আর 
{\bng EbNNG, {O}, Aar}
\\Interjection: 
% \bn হায়, ওহ, ওঃ, ওমা 
{\bng Hay, {O}H, {O}{h}, {O}ma}
\\Preposition: 
% \bn দ্বারা, দিয়ে, তবে
{\bng dWara, idey, teb}
\item \textbf{Emoji Remove :}
Emojis are used to convene text mode. Emojis have been eliminated for the sake of simplicity and simpler data. We will simply be focused on words in this paper only. Example: \\
% \bn আপনি একটা ঘোড়ার ডিম। 
{\bng Aapin EkTa egharhar iDm} \includegraphics[height=1em]{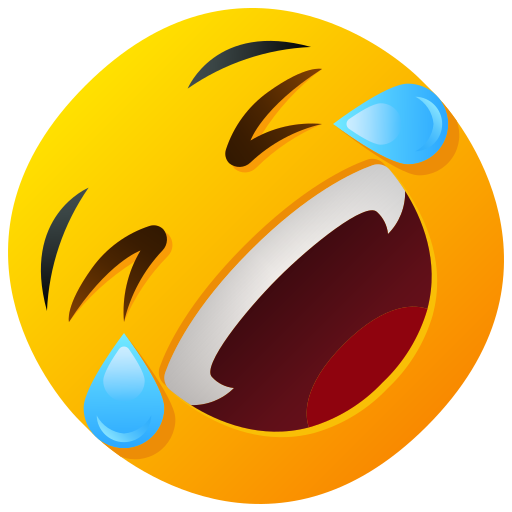} \includegraphics[height=1em]{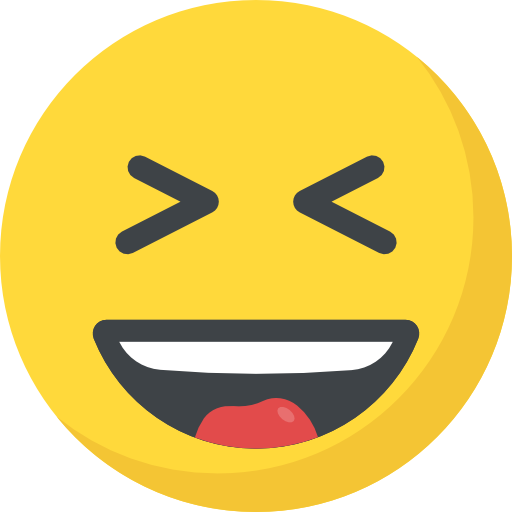} \includegraphics[height=1em]{emoji/laughter.png}
	 (Raw text) \\- 
                     % \bn আপনি একটা ঘোড়ার ডিম।
                     {\bng Aapin EkTa egharhar iDm}
                     (Processed Text)
\end{itemize}

\subsection{Data Distribution}
As we have mentioned there are four types of racist text. But unable to find an equal number of data in each class. So, our dataset is very imbalanced to the multi-class target labels. The data distribution in the \autoref{tab:data_distro}. 

\begin{table}[!h]
    \centering
    \caption{Data distribution}
    \label{tab:data_distro}
    \begin{tabular}{cc}
    \hline
        \textbf{Data Label} & \textbf{Data Count} \\
        \hline
        Representational Racism & 1974 \\
        
        Ideological Racism & 1062\\
        
        Discursive Racism & 1905\\
        
        Normal & 1214\\

         Total & 6155\\
        \hline
    \end{tabular}
\end{table}

\begin{figure*}[t]
    \label{fig:datasetConstruction}
  \centering

  \includegraphics[width=1.5\columnwidth]{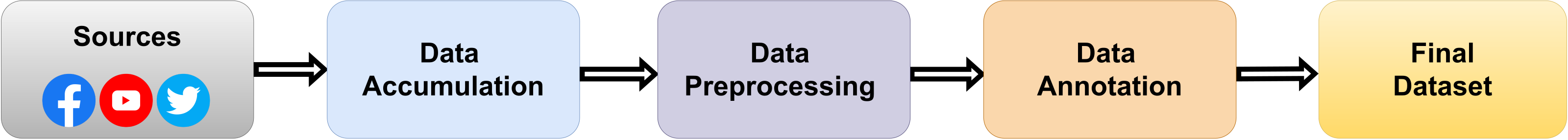}

  \caption{Dataset construction process}
\end{figure*}
\section{Methodology}
\begin{figure}[htbp]
\centering
\includegraphics[scale=0.5]{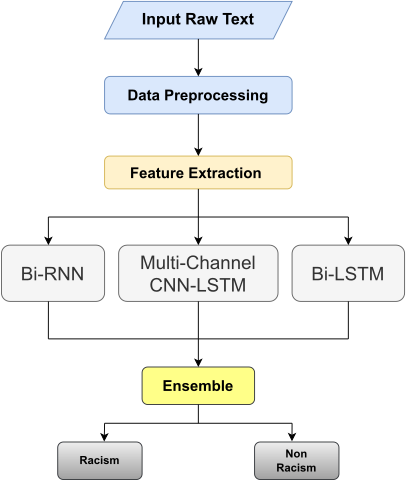}
\caption{A schematic diagram of our work.}
\label{fig: workflowDiagram}
\end{figure}

This section portrays the workflow of our research. \autoref{fig: workflowDiagram} displays the tactics we've suggested. At first, we collected data manually and preprocessed it. Word Embedding has been used for feature extraction. Finally, we have proposed deep learning models using MCNN-LSTM, LSTM \& RNN. 

\subsection{Feature Extraction:}
Word embedding, a crucial component in NLP, represents text by learning a representation where words with similar meanings share comparable features. This technique facilitates the training of machine learning models on textual data by extracting meaningful features. Word embeddings provide context to sentences. However, for the Bengali language, there is a scarcity of pre-trained BERT models compared to the abundance available for English. In this paper, \textbf{three} BERT pre-trained architectures are utilized for word embeddings. Table \ref{tab:bert} displays the number of embeddings provided by these BERT models.
\begin{table}[!h]
\caption{Different BERT model used for Embedding}
\label{tab:bert}
\begin{tabular}{cccc}
\hline
\textbf{BERT}           &  \begin{tabular}[c]{@{}c@{}}Bangla\\BERT\cite{bhattacharjee2021banglabert}\end{tabular} & \begin{tabular}[c]{@{}c@{}}BanglaBERT \\ Base\cite{Sagor_2020}\end{tabular} & sahajBERT\footnote{ \url{https://huggingface.co/neuropark/sahajBERT}} \\ \hline
\textbf{Embedding Size} & 768                                                           & 768       & 1024                                                                             \\ \hline
\end{tabular}
\end{table}

% \begin{enumerate}
%     \item csebuetnlp/banglabert\cite{bhattacharjee2021banglabert}
%     \begin{enumerate}
%         \item [!h]
%         They have transformed the sentence into 768 embeddings.
%     \end{enumerate}
%     \item sagorsarker/bangla-bert-base\cite{Sagor_2020}
%         \begin{enumerate}
%         \item [!h]
%         Here they transformed the sentence into 768 embeddings.
%     \end{enumerate}
%     \item neuropark/sahajBERT\footnote{ https://huggingface.co/neuropark/sahajBERT}
%         \begin{enumerate}
%         \item [!h]
%         But They have transformed the sentence into 1024 embeddings.
%     \end{enumerate}
% \end{enumerate}

\subsection{Model Architecture:}

\begin{figure*}[b]
\centering
\includegraphics[width=1.5\columnwidth]{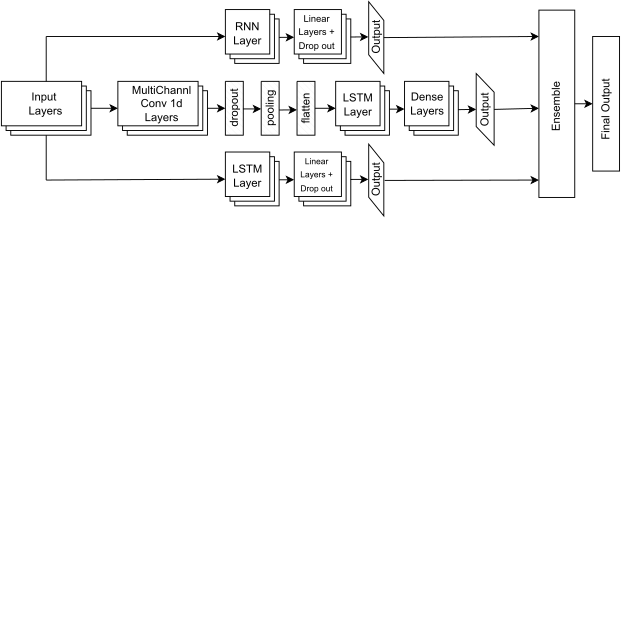}
\caption{Architecture of our proposed MCNN-LSTM.}
\label{fig: modelArchitecture}
\end{figure*}

We have proposed an Ensemble Deep Learning Model approach to classify target labels. Here, \autoref{fig: modelArchitecture} is an overview of our model. We have used Multi-Channel CNN-LSTM, Bi-LSTM, and Bi-RNN. The model is built using TensorFlow \& PyTorch.

\subsection{Fine Tuned Hyper-parameters:}
We did fine-tune the model. We have trained and tested with all three pertained word embedding models individually. \autoref{tab: inputSize} shows the input size of three models. Along with word-embedding models neural network model has been varied for RNN and LSTM, we have a single input for all three types of embeddings. Conversely, the MCNN-LSTM architecture necessitates the utilization of three parallel inputs, aligning with the design of its three-channel framework. \autoref{tab:hyperparameters} shows all the necessary hyperparameters of the models.
\begin{table}
    \centering
    \caption{Input size of different models}
    \label{tab: inputSize}
    \begin{tabular}{p{3cm}|c|p{3cm}}
        \hline
        \textbf{Word-Embedding} & \textbf{Model} & \textbf{Input Size} \\
        \hline
                & &\\

        \multirow{8}{3cm}{Bangla Bert, Bangla Bert Base, Sahaj Bert} 
        & RNN & 768, 768, 1024 \\
        & &\\

        \cline{2-3}
                & &\\

         & LSTM & 768, 768, 1024\\
                 & &\\

        \cline{2-3}
                         & &\\

        & & (768, 768, 768),\\
         & MCNN-LSTM &  (768, 768, 768), \\
        & & (1024, 1024, 1024)\\
         
        \hline
    \end{tabular}
\end{table}

\begin{table}[!h]
    \centering
    \caption{Hyperparameters used in explored models}
    \begin{tabular}{cc}
       \hline
        Kernel Size (MCNN-LSTM) & (4, 6, 8) \\
        
        Activation Function & Sigmoid\\
        
         Optimizer & Adam\\
         
         Epochs & 10, 18\\
        
         Batch Size & 10 \\
         
         Learning Rate & 0.0001\\
         
        Trian-Test Split & 80:20\\
         
          Loss Function & Cross Entropy Loss\\
         \hline
    \end{tabular}
    \label{tab:hyperparameters}
\end{table}

\subsection{Ensemble Approach}

The ensemble approach is basically a combined decision from all three models that we have developed. Firstly, we read our Corpora Racism dataset. After that, the dataset was split into Train-Test. Then, with the test dataset, we predicted the output for all three models individually. Finally, all three prediction labels from the three models have been averaged into one ensemble prediction. Thus, we have our final model output. 

% \begin{algorithm}
% \caption{Ensemble Approach}
% \label{algo:ensembleApproach}
% \KwData{Load binary dataset from \texttt{embeddedDatasetPath}}
% \KwResult{Display the results}

% \SetKwInOut{Input}{Input}
% \SetKwInOut{Output}{Output}

% \Input{$X\_train$, $X\_test$, $y\_train$, $y\_test$}
% \Output{$predicted\_labels$}

% Load binary dataset from \texttt{embeddedDatasetPath}\;
% Split the dataset into training and testing sets\;
% \quad $X\_train$, $X\_test$, $y\_train$, $y\_test \gets$ \texttt{trainTestSplit(datasetBinary, test\_size=0.2)}\;

% Make predictions using the RNN model\;
% \quad $output1 \gets$ \texttt{model\_RNN.predict($X\_test$)}\;

% Make predictions using the LSTM model\;
% \quad $output2 \gets$ \texttt{model\_LSTM.predict($X\_test$)}\;

% Make predictions using the MCNN-LSTM model\;
% \quad $output3 \gets$ \texttt{model\_MCNNLSTM.predict([$X\_test$, $X\_test$, $X\_test$])}\;

% Calculate ensemble predictions by averaging\;
% \quad $ensemble\_predictions \gets \texttt{Average}($output3$, $output2$, $output1$)$\;

% Set the threshold for binary classification (e.g., 0.5)\;

% Generate predicted labels by applying the threshold\;
% \quad $predicted\_labels \gets (\texttt{ensemble\_predictions} > \texttt{threshold})$\;

% Display the results\;
% \quad \texttt{DisplayResult}($y\_test$, $predicted\_labels$)\;

% \end{algorithm}

\section{Experimental Results}
\subsection{Experiments}
Several experiments were conducted with the model, and Bi-RNN, Bi-LSTM, and MCNN-LSTM showed outstanding performance. Different optimizers, including Adam, Nadam, and Radam, were employed, resulting in noticeable progress. Custom-built models were fine-tuned, with both LSTM and RNN models performing well in text classification. The convolution technique, particularly with the LSTM model, demonstrated superior performance. The Multi-Channel CNN-LSTM achieved an accuracy of \textbf{86.93\%} with Sahaj BERT Embeddings, and the ensemble approach reached the highest accuracy of \textbf{87.94\%}.

\subsection{Evaluation Metrics}
Performance is assessed using a variety of conventional assessment metrics, including Accuracy, and F1-Score which are obtained from precision and recall considerations.
% Model Uncertainty, and Cohen's Kappa score. These metrics allow for the quantification and comparison of performance results. \autoref{eq:uncertainity} formula is used to calculate predictive entropy which portrays the model's uncertainty. Here, $ \overline{P_i}
% $ is the mean probability of class \textit{i} across all predictions. $\epsilon$ is a small positive constant. 
% \begin{equation}
%     \label{eq:uncertainity}
% H = - \sum_i (\overline{P_i} \cdot \log(\overline{P_i} + \epsilon)) 
% \end{equation}

% Cohen's Kappa score is measured using \autoref{eq:cohenKappa}. Basically, Kappa($\kappa$)\footnote{$https://en.wikipedia.org/wiki/Cohen\%27s_kappa$} is a more precise version of Accuracy metrics. All these metrics are implemented with the help of Sci-Kit Learn.

% \begin{equation}
% \label{eq:cohenKappa}
% \kappa = (p_o - p_e) / (1 - p_o)
% \end{equation}

\subsection{Experimental Results}

\begin{figure*}[]
    \centering
    \includegraphics[scale=0.2]{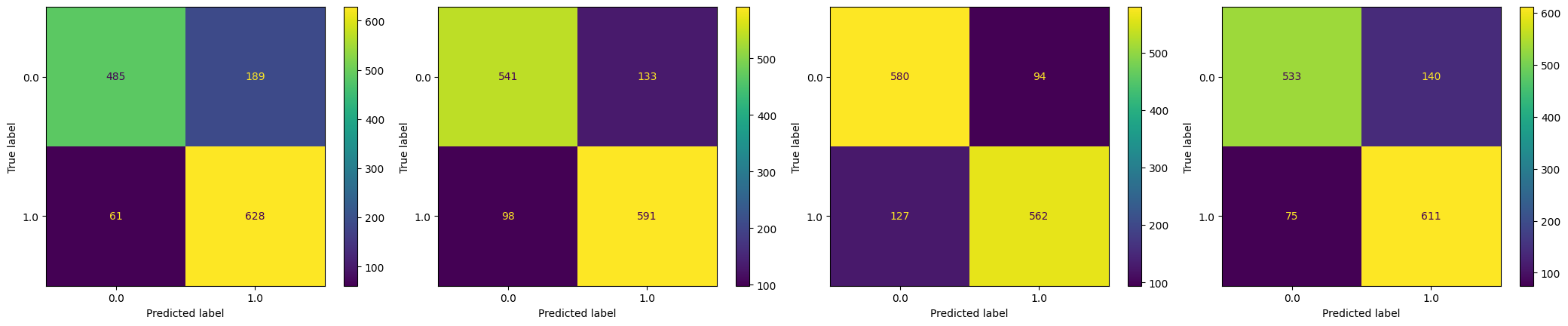}
    \caption{Confusion matrix of Bi-RNN, Bi-LSTM, MCNN-LSTM, and Ensemble Models with the embeddings from Sahaj BERT.(\textit{left-right})}
    \label{fig:confusion_matrix}
\end{figure*}

\begin{figure*}[]
    \centering

    \includegraphics[scale=0.13]{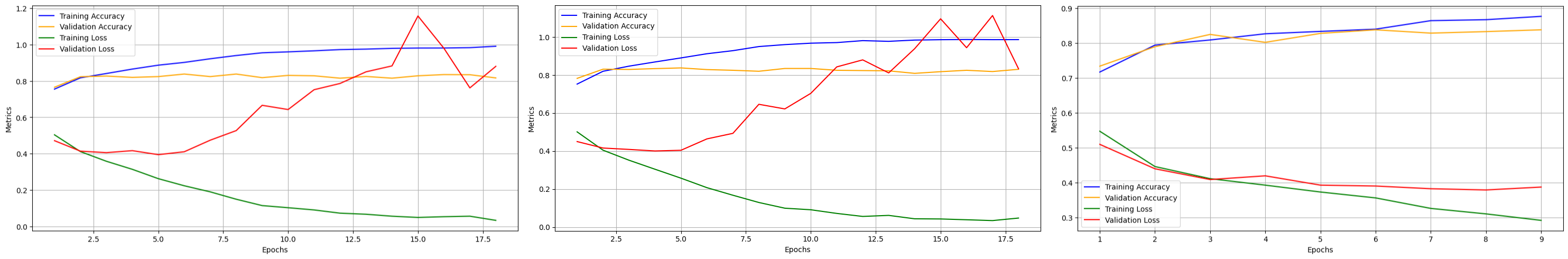}

    \caption{Accuracy Loss plot of Bi-RNN, Bi-LSTM, and MCNN-LSTM Models with the embeddings from Bangla Bert Base. (\textit{left-right})}

    \label{fig:accuracyLossPlot}
\end{figure*}

This section portrays the best results from all the experiment that has been conducted. The evaluation metrics, namely Precision, Recall, Accuracy, and F1-Score have been presented in  \autoref{tab:performance_table}.\\
Also, we have shown the Confusion Matrix for all the models in \autoref{fig:confusion_matrix} and the Accuracy-Loss plot for all three models in \autoref{fig:accuracyLossPlot}.

\begin{table}[!h]
\centering
\caption{Performance of the models. Class 0 and 1 denote Non-Racism and Racism respectively.}
\begin{tabular}{ccccccc}
\hline

            \multirow{3}{*}{\begin{tabular}[c]{@{}c@{}}\textbf{Word} \\\textbf{Embedd-}\\\textbf{ings}\end{tabular}} & \multirow{3}{*}{\textbf{Model}} & \multirow{3}{*}{\textbf{Class}} & \multicolumn{3} {c} {\textbf{Metrics}}  & \multirow{3}{4em}{\textbf{Acc(\%})}\\
\cline{4-6}\\
            {} & {} & {} & {\textbf{P}} & {\textbf{R}} & {\textbf{F1}} & {} \\
\hline

             \multirow{8}{*}{{\begin{tabular}[c]{@{}c@{}}Bangla\\BERT\end{tabular}}} & \multirow{2}{4em}{Bi-RNN} & 1 & 0.80 & 0.86 & 0.83 &  \multirow{2}{*}{82.56} \\
\cline{3-6}
             {} & {} & 0 & {0.85} & {0.78} & {0.81} & \\
\cline{2-7}

             {} & \multirow{2}{*}{Bi-LSTM} & 1 & {0.81} & {0.87} & {0.84} & \multirow{2}{*}{83.63} \\
\cline{3-6}
             {} & {} & 0 & {0.86} & {0.80} & {0.83} &  \\
\cline{2-7}

             {} & \multirow{2}{*}{{\begin{tabular}[c]{@{}c@{}}MCNN-\\LSTM\end{tabular}}} & 1 & {0.82} & {0.91} & {0.86} & \multirow{2}{*}{85.25} \\
\cline{3-6}
             {} & {} & 0 & {0.90} & {0.79} & {0.84} &  \\
\cline{2-7}
 
             {} & \multirow{2}{*}{Ensemble} & 1 & {0.83} & {0.87} & {0.85} & \multirow{2}{*}{85.27} \\
\cline{3-6}
             {} & {} & 0 & {0.86} & {0.82} & {0.84} &\\
 \cline{2-7}
    \hline

             \multirow{8}{*}{{\begin{tabular}[c]{@{}c@{}}Bangla\\BERT\\Base\end{tabular}}} & \multirow{2}{4em}{Bi-RNN} & 1 & 0.77 & 0.91 & 0.83 & \multirow{2}{*}{82.61} \\
\cline{3-6}
             {} & {} & 0 & {0.89} & {0.72} & {0.80} & \\
\cline{2-7}

             {} & \multirow{2}{*}{Bi-LSTM} & 1 & {0.86} & {0.82} & {0.84} & \multirow{2}{*}{84.13} \\
\cline{3-6}
             {} & {} & 0 & {0.82} & {0.86} & {0.84} &  \\
\cline{2-7}

             {} & \multirow{2}{*}{{\begin{tabular}[c]{@{}c@{}}MCNN-\\LSTM\end{tabular}}} & 1 & {0.86} & {0.82} & {0.84} & \multirow{2}{*}{84.26} \\
\cline{3-6}
             {} & {} & 0 & {0.82} & {0.86} & {0.84} & \\
\cline{2-7}
 
             {} & \multirow{2}{*}{Ensemble} & 1 & {0.81} & {0.89} & {0.85} & \multirow{2}{*}{84.51} \\
\cline{3-6}
             {} & {} & 0 & {0.88} & {0.79} & {0.83} &  \\
 \cline{2-7}
    \hline

             \multirow{8}{*}{{\begin{tabular}[c]{@{}c@{}}sahaj\\ BERT\end{tabular}}} & \multirow{2}{4em}{Bi-RNN} & 1 & 0.83 & 0.89 & 0.86 & \multirow{2}{*}{85.03}\\
\cline{3-6}
             {} & {} & 0 & {0.88} & {0.81} & {0.84} &\\
\cline{2-7}

             {} & \multirow{2}{*}{Bi-LSTM} & 1 & {0.90} & {0.72} & {0.80} & \multirow{2}{*}{82.15} \\
\cline{3-6}
             {} & {} & 0 & {0.76} & {0.92} & {0.83} & \\
\cline{2-7}

             {} & \multirow{2}{*}{\begin{tabular}[c]{@{}c@{}}MCNN-\\LSTM\end{tabular}} & 1 & {0.79} & {0.93} & {0.85} & \multirow{2}{*}{86.93} \\
\cline{3-6}
             {} & {} & 0 & {0.91} & {0.74} & {0.82} & \\
\cline{2-7}
 
             {} & \multirow{2}{*}{Ensemble} & 1 & {0.83} & {0.94} & {0.88} & \multirow{2}{*}{\textbf{87.94}} \\
\cline{3-6}
             {} & {} & 0 & {0.93} & {0.80} & {0.86} & \\
 \cline{2-7}
    \hline
    \end{tabular}
    \label{tab:performance_table}
\end{table}

\subsection{Performance Comparison}
Several works are available on detecting racist text in other languages \& hate speech detection. But in the Bengali language, there is no work on racism particularly. Racism is a very precise subdomain of Hate Speech. For comparison, we have exhibited two papers in \autoref{tab:otherPaperCompare}.

\begin{table}[!h]
\caption{Comparison of our work to others.}
\begin{tabular}{ccp{1.5cm}c}
\hline
Paper Title     & Approach                                                    & Dataset & Result                                                             \\\hline
Andrades et al.\cite{benitez2022detecting} & BERT                                                        &    HaterNet, HatEval, PHARM \cite{pereira2019detecting, i2019multilingual, vrysis2021web}     & \begin{tabular}[c]{@{}c@{}}Precision-\\ .8522\end{tabular}         \\ \hline
Mozafari et al.\cite{mozafari2020bert} & BERT                                                        &   Waseem and Hovy's Dataset \cite{waseem2016hateful}      & \begin{tabular}[c]{@{}c@{}}F1-Score- \\ .88 \& \\ .92\end{tabular} \\ \hline
Ours            & \begin{tabular}[c]{@{}c@{}}BERT + \\ MCNN-LSTM\end{tabular} &    Corpora Racism \footnote{\url{https://github.com/sjalim/Detecting-Racist-Text-in-Bengali-An-Ensemble-Deep-Learning-Framework}}     & \begin{tabular}[c]{@{}c@{}}Accuracy-\\ 87.94\%\end{tabular} \\ \hline      
\end{tabular}
\label{tab:otherPaperCompare}
\end{table}

\subsection{Result Analysis}
The ensemble approach got the highest accuracy with Sahaj BERT embeddings. 
MCNN-LSTM performed comparatively better with respect to LSTM and RNN models. As we know BERT extracts the context of a sentence to some embeddings. After that embeddings are fed to the models. In that case, LSTM and RNN models are well-known for categorizing sequential data in the text analysis domain. But surprisingly we see better results with MCNN-LSTM. MCNN-LSTM is a combination of the convolution technique and the LSTM model. Convolution Neural Networks compress the information into a smaller size with the help of different convolution approaches. In our case, different kernel size has been used for convolution.\\
Another factor is observed that depending on different word embeddings the result verifies. From \autoref{tab:hyperparameters} we have seen two types of embedding size 768 \& 1024. Analysing the result it is clear that \textbf{ Sahaj BERT embeddings} are sustaining a more proper context of a text for our dataset.\\
Moreover, classwise F1-score are racsim \textbf{0.88} \& \textbf{0.86} for ensemble approach. Thus ensemble approach is quite helpful in increasing performance.

\section{Conclusion and Future Work}

Though we have successfully detected the racist. Due to a lack of data, we have been unable to categorize the dataset to multi-class with formidable accuracy. On the other hand, our models performed quite well with binary classification. Ensemble model architecture pushed the performance a little bit higher. The ensemble approach has the highest accuracy 87.94\%. In all experimental trials, it can be observed that the MCNN-LSTM model consistently outperformed both the RNN and LSTM models. In this paper, we have focused on racist text excluding all the generic hate speech comments. This work can be extended by increasing the dataset. We intend to increase our dataset and make a more precise model in the near future.

% \bibliographystyle{biblography/IEEEtran}
% \bibliography{biblography/IEEEexample}

\end{document}